\documentclass[10pt,twocolumn,letterpaper]{article}

\usepackage{cvpr}
\usepackage{times}
\usepackage{epsfig}
\usepackage{graphicx}
\usepackage{amsmath}
\usepackage{amssymb}
\usepackage{bm}
\usepackage{color}
\usepackage{mathtools}
\usepackage{array, booktabs, makecell}
\usepackage{siunitx, mhchem}
\usepackage{xcolor}
\definecolor{saddle-brown}{RGB}{139, 69, 19}
\definecolor{dark-green}{RGB}{0,125, 0}
\definecolor{light-green}{RGB}{0,200, 0}
\definecolor{coolgrey}{rgb}{0.55, 0.57, 0.67}
\definecolor{darkelectricblue}{rgb}{0.33, 0.41, 0.47}

\usepackage{booktabs}
\newcommand{\ra}[1]{\renewcommand{\arraystretch}{#1}}
\long\def\ignorethis#1{}

\newcommand\blfootnote[1]{%
	\begingroup
	\renewcommand\thefootnote{}\footnote{#1}
	\addtocounter{footnote}{-1}
	\endgroup
}

\newbox\jsavebox
\newcommand{\jsubfig}[2]{%
	\sbox\jsavebox{#1}%
	\parbox[t]{\wd\jsavebox}{\centering\usebox\jsavebox\\#2}%
}

\usepackage[pagebackref=true,breaklinks=true,letterpaper=true,colorlinks,bookmarks=false]{hyperref}

\cvprfinalcopy 

\def\cvprPaperID{5} 

\ifcvprfinal\fi
\begin{document}
	
	\title{READ: Recursive Autoencoders for Document Layout Generation}


\author{Akshay Gadi Patil$^{\dagger}$
\\
Simon Fraser University
\and
Omri Ben-Eliezer$^{\dagger}$
\\
Tel-Aviv University
\and
Or Perel\\
Amazon
\and
Hadar Averbuch-Elor$^{\mathsection}$ \\
Cornell Tech, Cornell University
}

	\maketitle
	\thispagestyle{empty}

	\newcommand{\aksh}[1]{{\color{blue}Akshay: #1}}
	\newcommand{\he}[1]{{\color{red}Hadar: #1}}
	\newcommand{\todo}[1]{{\color{red}TODO: #1}}
	\newcommand{\fillit}[1]{{\color{red}#1}}
	\newcommand{\xd}[1]{{\color{green}Omri: #1}}
	\newcommand{\orp}[1]{{\color{blue}Or: #1}}
	\newcommand{\rev}[1]{{\color{black}#1}}
	
	\begin{abstract}
Layout is a fundamental component of any graphic design. 
Creating large varieties of plausible document layouts can be a tedious task, requiring numerous constraints to be satisfied, including local ones relating different semantic elements and global constraints on the general appearance and spacing. 
In this paper, we present a novel framework, coined READ, for REcursive Autoencoders for Document layout generation, to generate plausible 2D layouts of documents
in large quantities and varieties. First, we devise an exploratory recursive method to extract a structural decomposition of a single document.
%
Leveraging a dataset of documents annotated with labeled bounding boxes, our recursive neural network learns to map the structural representation, given in the form of a simple hierarchy, to a compact code, the space of which is approximated by a Gaussian distribution. Novel hierarchies can be sampled from this space, obtaining new document layouts.
Moreover, we introduce a combinatorial metric to measure structural similarity among document layouts. We deploy it to show that our method is able to generate highly variable and realistic layouts.
We further demonstrate the utility of our generated layouts in the context of standard detection tasks on documents, showing that detection performance improves when the training data is augmented with generated documents whose layouts are produced by READ.

\end{abstract}

	\blfootnote{$^{\dagger}$ work done as an Intern at Amazon}
\blfootnote{$^{\mathsection}$ work done while working at Amazon}

\section{Introduction}
\label{sec:intro}
\emph{``Do not read so much, look about you and think of what you see there."} \qquad \qquad \qquad \qquad \qquad \emph{-Richard Feynman}\\

\begin{figure}
    \centering
   \jsubfig{\fbox{\includegraphics[height=3.5cm]{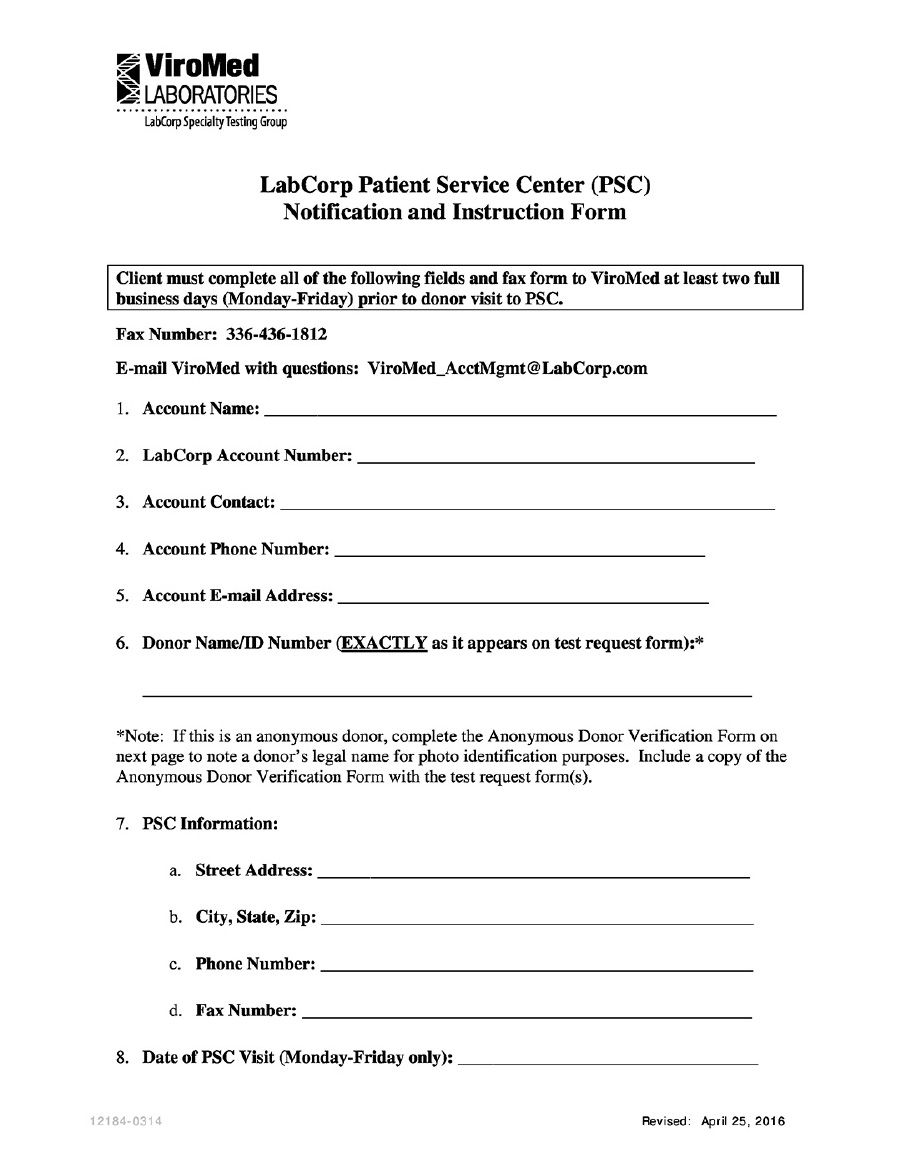}}}
	{Real document}%
 	\hfill%
\jsubfig{\fbox{\includegraphics[height=3.5cm]{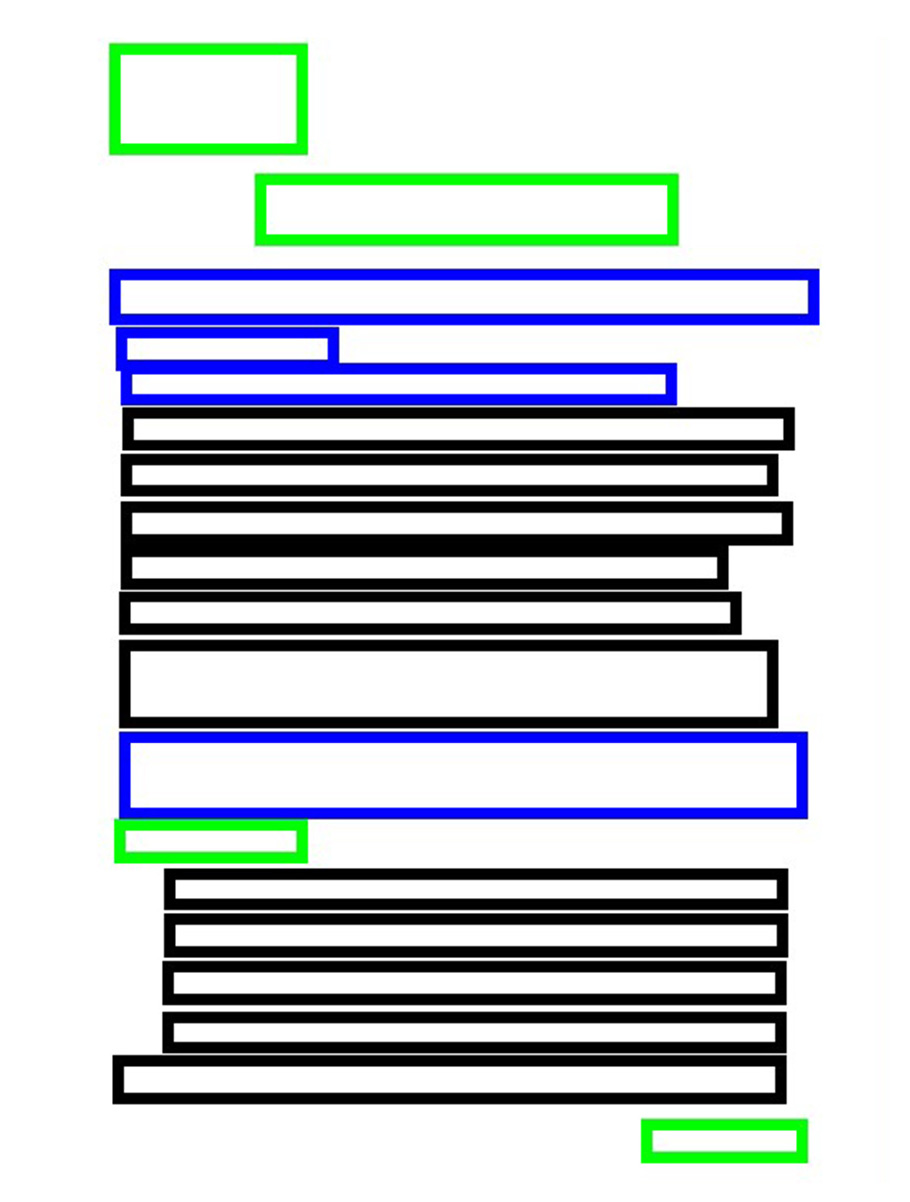}}}
	{Real layout}%
	\hfill%
\jsubfig{\fbox{\includegraphics[height=3.5cm]{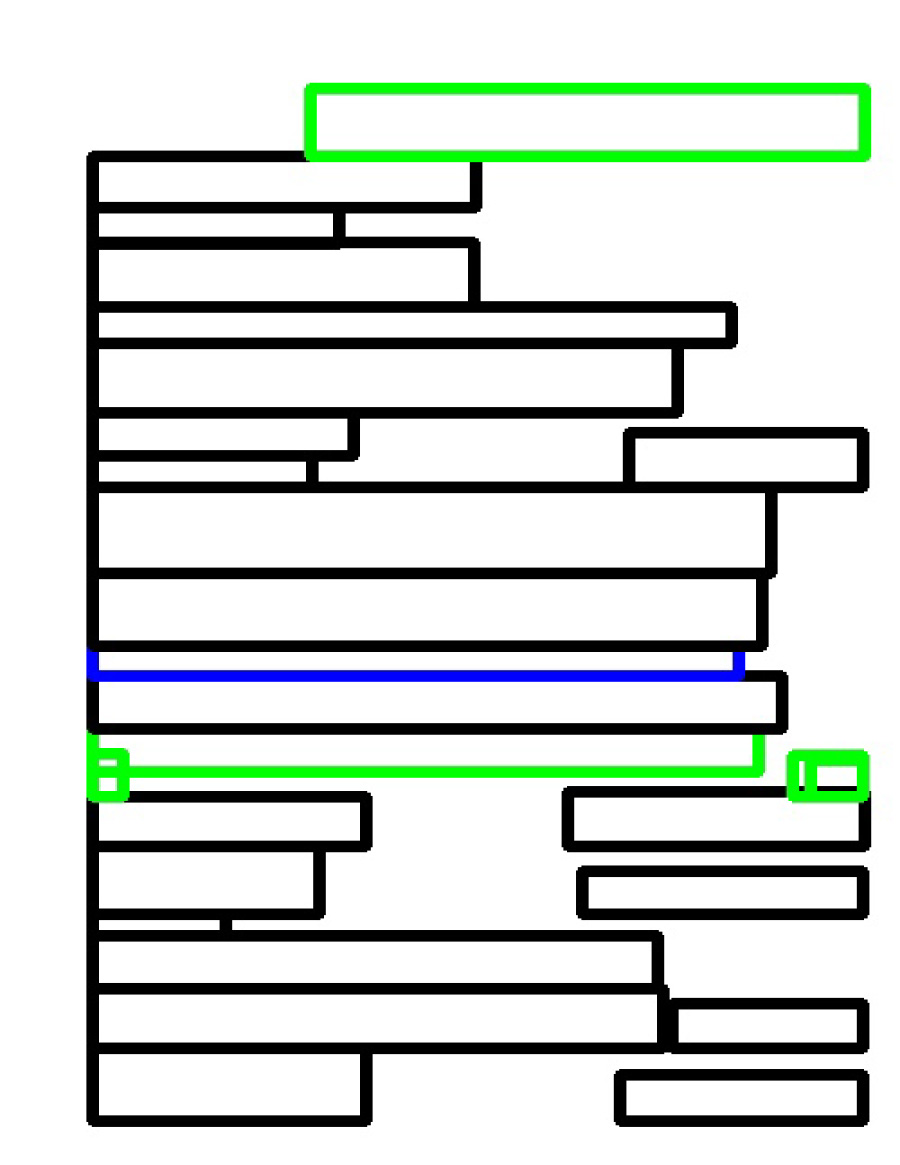}}}
	{Generated layout}%
	\vspace{8pt}
    \caption{
    Given a collection of training examples -- annotated layouts (middle) of real-world documents (such as the fillable form on the left) -- our method generates synthetic layouts (right) resembling those in the training data. 
    Semantically labeled regions are marked in unique colors.
    }
    \label{fig:teaser}
    \vspace{-1 em}
\end{figure}

Layouts are essential for effective communication and targeting one's visual attention. From newspapers articles, to magazines, academic manuscripts, websites and various other document forms, layout design spans a plethora of real world document categories and receives the foremost editorial consideration. However, while the last few years have experienced growing interests among the research community in generating novel samples of images \cite{karras2018style,oord2016pixel}, audio \cite{oord2016wavenet} and 3D content \cite{li2017grass,li2018grains,Wang2018Scene,wu2016learning}, little attention has been devoted towards automatic generation of large varieties of plausible document layouts. To synthesize novel layouts, two fundamental questions must first be addressed. What is an appropriate representation for document layouts? And how to synthesize a new layout, given the aforementioned representation? 

The first work to explicitly address these questions is the very recent LayoutGAN of Li et al. \cite{li2019layoutgan}, which approaches layout generation using a generative adversarial network (GAN) \cite{goodfellow2014generative}. They demonstrate impressive results in synthesizing plausible document layouts with up to nine elements, represented as bounding boxes in a document. However, various types of highly structured documents can have a substantially higher number of elements -- up to tens or even hundreds.\footnote{As an example, consider the popular US tax form 1040; See \url{https://www.irs.gov/pub/irs-pdf/f1040.pdf}.} Furthermore, their training data constitutes about $25$k annotated documents, which may be difficult to obtain for various types of documents. Two natural questions therefore arise: Can one devise a generative method to synthesize highly structured layouts with a large number of entities? And is it possible to generate synthetic document layouts without requiring a lot of training data? 
\begin{figure*}[th]
 \centering
    \includegraphics[width=0.90\textwidth] {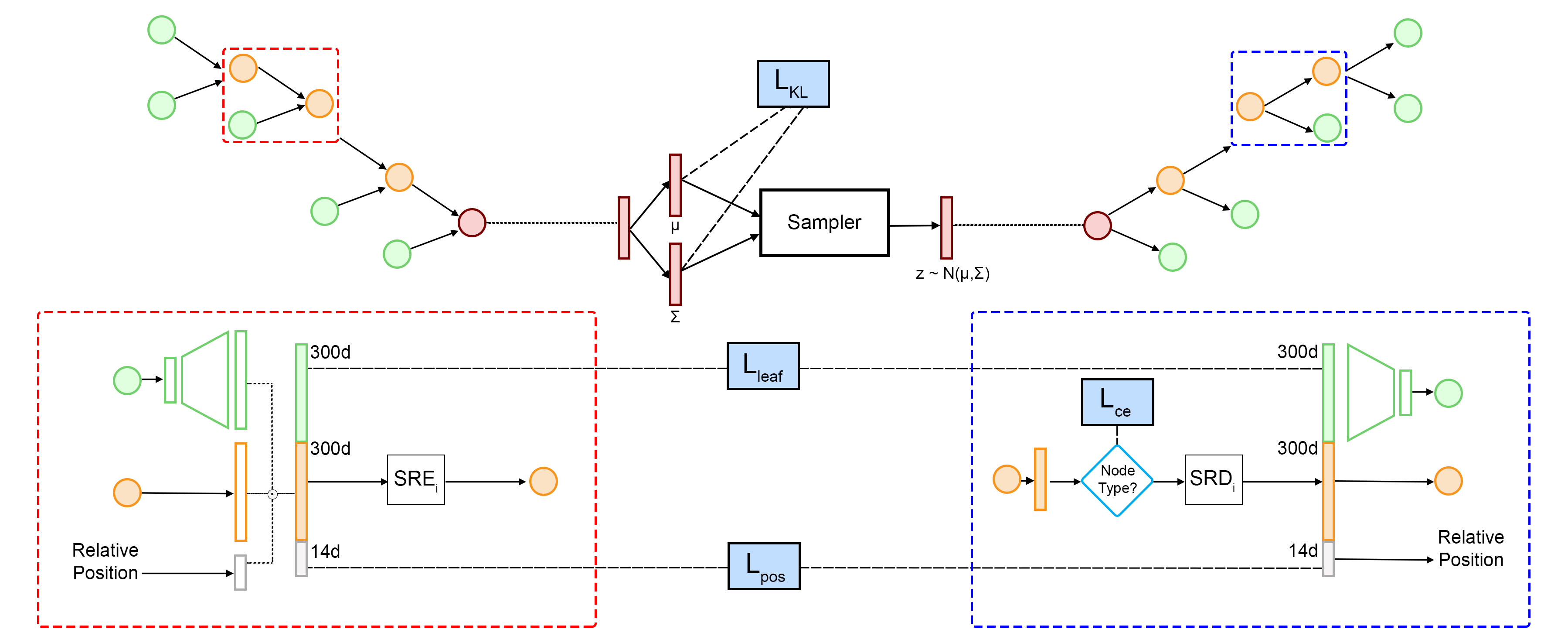}
    \caption{Overview of our RvNN-VAE framework. Training hierarchies are constructed for every document in the dataset. These hierarchies are mapped to a compact code (in a recursive fashion according to the encoder network marked in red), the space of which is approximated by a Gaussian distribution. 
    Novel hierarchies can be sampled from this space (and decoded recursively according to the decoder network marked in blue), obtaining new document layouts.   }
    \label{fig:pipeline}
\end{figure*}
In this work, we answer both questions affirmatively.
Structured hierarchies are natural and coherent with human understanding of document layouts.
We thus present \emph{READ}: a \emph{generative} \emph{recursive} neural network (RvNN) that can appropriately model such structured data. Our method enables generating large quantities of plausible layouts containing dense and highly variable groups of entities, using just a few hundreds of annotated documents. With our approach, a new document layout can be generated from a random vector drawn from a Gaussian in a fraction of a second, following the pipeline shown in Figure \ref{fig:pipeline}.

Given a dataset of annotated documents, where a single document is composed of a set of labeled bounding boxes, we first construct document hierarchies, which are built upon connectivity and implicit symmetry of its semantic elements. These hierarchies, or trees, are mapped to a compact code representation, in a recursive bottom-up fashion.
The resulting fixed length codes, encoding trees of different lengths, are constrained to roughly follow a Gaussian distribution by training a Variational Autoencoder (VAE). A novel document layout can be generated by a recursive decoder network that maps a randomly sampled code from the learned distribution, to a full document hierarchy. 
To evaluate our generated layouts, we introduce a new combinatorial metric (\texttt{DocSim}) for measuring layout similarity among structured multi-dimensional entities, with documents as a prime example. We use the proposed metric to show that our method is able to generate layouts that are representative of the latent distribution of documents which it was trained on. As one of the main motivations to study synthetic data generation methods stems from their usefulness as training data for deep neural networks, we also consider a standard document analysis task. We augment the available training data with synthetically generated documents whose layouts are produced by READ, and demonstrate that our augmentation boosts the performance of the network for the aforementioned document analysis task.

	\section{Related Work}
Analysis of structural properties and relations between entities in documents is a fundamental challenge in the field of information retrieval. While local tasks, like optical character recognition (OCR) have been addressed with very high accuracy, the global and highly variable nature of document layouts has made their analysis somewhat more elusive.
Earlier works on structural document analysis mostly relied on various types of specifically tailored methods and heuristics (e.g., \cite{Baird1992StructuredDI,Breuel03highperformance,Kasturi2002,Ogorman1993}
Recent works have shown that deep learning based approaches significantly improve the quality of the analysis; e.g., see the work of Yang et al. \cite{yang2017learning}, which uses a joint textual and visual representation, viewing the layout analysis as a pixel-wise segmentation task.
Such modern deep learning based approaches typically require a large amount of high-quality training data, which call for suitable methods to synthetically generate documents with real-looking layout \cite{li2019layoutgan} and content \cite{DesignSemantics2018}. Our work continues the line of research on synthetic layout generation, showing that our synthetic data can be useful to augment training data for document analysis tasks.

Maintaining reliable representation of layouts has shown to be useful in various graphical design contexts, which typically involve highly structured and content-rich objects. The most related work to ours is the very recent LayoutGAN of Li et al. \cite{li2019layoutgan}, which aims to generate realistic document layouts using a generative adversarial networks (GAN) with a wireframe rendering layer. Zheng et al.  \cite{zheng2019content} also employ a GAN-based framework in generating documents, however, their work focuses mainly on content-aware generation, using the content of the document as an additional prior. Unlike Convolutional Neural Networks (CNNs) that operate on large dimensional vectors and involve multiple multi-channel transformations, in our work,
we use recursive neural networks, which operate on low-dimensional vectors and employ two-layer perceptrons to merge any two vectors. Hence, they are computationally cheaper, plus can learn from just a few training samples.

Deka et al. \cite{deka2017rico} use an autoencoder to perform layout similarity search to simplify UI design for mobile applications. Ritchie et al. \cite{Ritchie2011Dtour} present a design exploration tool for layout and content based retrieval of similarly looking web pages. O'Donovan et al. \cite{odonovan2014} 
present an interactive energy-based model that allows novice designers to improve their page layout design. Swearngin et al. \cite{Swearngin2018Rewire} apply layout analysis to allow designers to manipulate layouts obtained from screenshots. More fundamentally, Talton et al. \cite{Talton2012DesignPatterns} leverage learned visual-structural and textual patterns learned from the data to obtain a formal grammar allowing to probabilistically generate new, similarly looking entities.


Recursive neural networks (RvNN) were first introduced by Socher et al. \cite{socher2011parsing,socher2013recursive} for parsing natural scenes and natural language sentences. Socher et al. \cite{socher2014recursive} comprehensively present applications of RvNNs for various tasks in computer vision. However, RvNNs did not enjoy as much attention as CNNs, until recently, when RvNNs coupled with generative models were shown to work effectively on previously unexplored paradigms such as generating 3D shape structures \cite{li2017grass,zhu2018scores} and indoor 3D scenes \cite{li2018grains}. Document layouts structurally resemble 3D indoor-scenes, in the sense
that semantic entities are loosely related and not bound by geometric connectivity (like parts in a 3D shape). But unlike indoor scenes, where any permutation of valid \emph{subscene} arrangements would synthesize plausible global scenes \cite{ma2017thesis,xu2014organizing}, semantic entities in a document must be placed at the right positions for the generated layout to look realistic; e.g., \emph{title} should always
appear at the top. In other words, document layouts enforce more global constraints.
	\section{Method}
\label{sec:method}

Our RvNN-VAE framework of generating layouts is trained on a dataset of documents with semantic-based labels. That is, each document is composed of a set of labeled bounding boxes (ex., magazine-articles are labeled with \emph{title}, \emph{paragraph}, and so on). 
We use the set of labeled bounding boxes, which we call the \emph{atomic units}, to build a training hierarchy for each document in our training set. 
These hierarchies are fed into our RvNN-VAE framework (see Figure \ref{fig:pipeline}) with a suitable training objective. 
Once trained, the RvNN-VAE network is used to generate a new layout by decoding a randomly sampled vector into a hierarchy of 2D bounding boxes with their corresponding semantic labels. 
\begin{figure}
    \centering
    \includegraphics[width=\linewidth, height = 0.6 \linewidth]{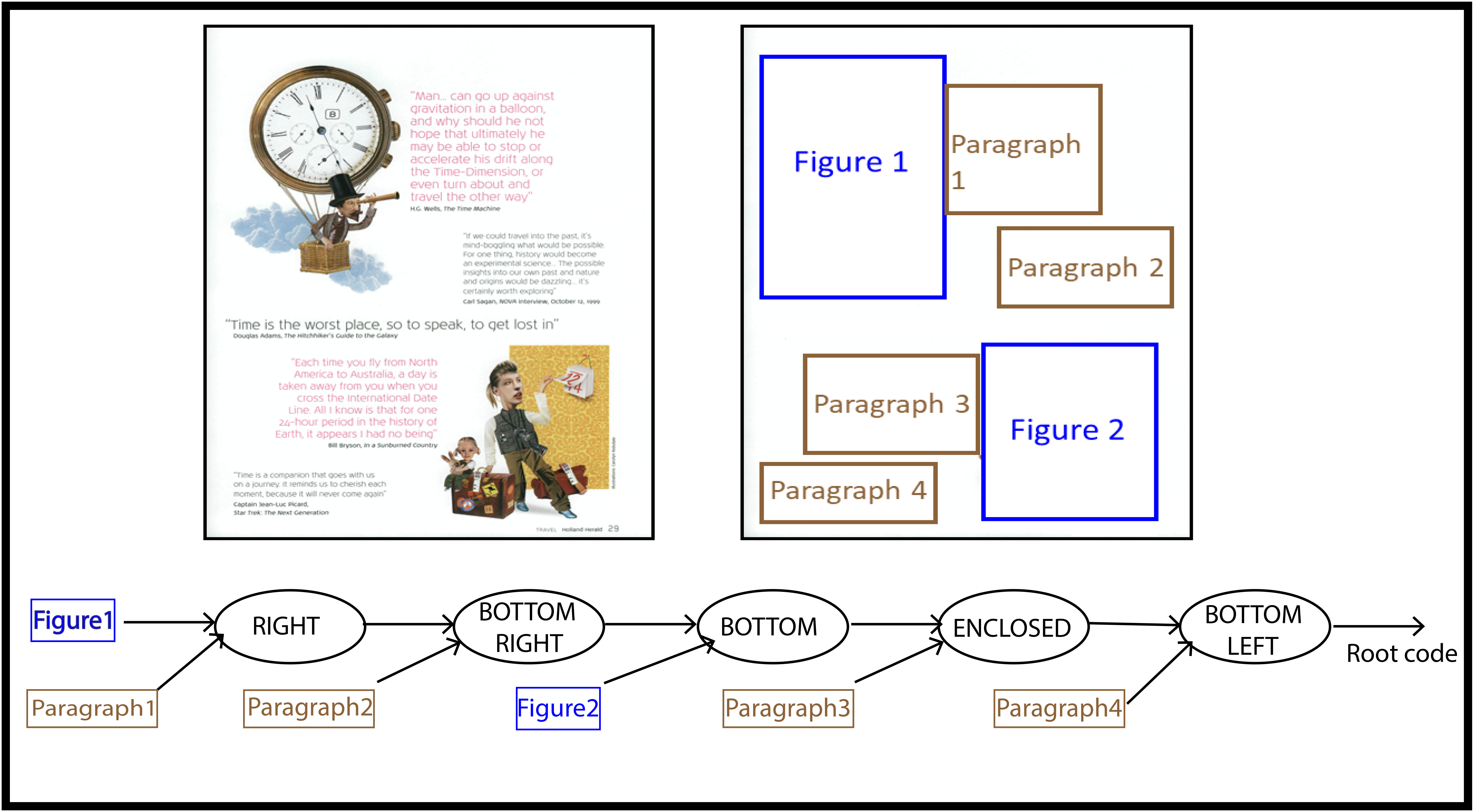}
    \caption{Exploratory layout extraction of a document from the ICDAR2015 \protect\cite{icdar2015} training set. The input document and the annotated boxes are shown on top. Note that when two boxes are merged, the merged bounding box is the union of the two boxes.}
    \label{fig:example_hierarchy}
    \vspace{-1.5 em}
\end{figure}


\subsection{Building training hierarchies}
\label{sec:building_hierarchies}

Given labeled bounding box annotations, we first extract a structural decomposition for every document in the training set, based on connectivity and \emph{implicit} symmetry of the atomic unit bounding boxes, by scanning the document from left-to-right and top-to-bottom. The results are stored as binary trees. 
We combine each pair of atomic elements, which we view as leaf nodes, into a union of boxes, viewed as an internal node, in a recursive manner, according to the relative position between the boxes. Internal nodes are also handled in a similar fashion. This exploratory process continues until all boxes are merged under a single root node. 
Figure \ref{fig:example_hierarchy} demonstrates the result of such an exploratory process on a single training sample. As the figure illustrates, we employ various types of spatial relationships (see Figure \ref{fig:7_types_of_spatial_relations}). 

As documents are designed by humans, there is a weak symmetric structure between related atomic unit boxes; fields that are spatially-related usually have similar box geometry. 
Traversing left-to-right and top-to-bottom does not always guarantee that atomic units with similar geometry are grouped together, e.g., boxes that are placed one below the other with the same box geometry may not be grouped together. However, we demonstrate that our RvNN-VAE framework is able to effectively capture relationships among the boxes with our simple traversal strategy, without any complex hand-crafted heuristics.



\subsection{Recursive model for document layouts}
\label{sec:rec}
Every atomic unit in the extracted hierarchies, to be used for training, is initially represented using its bounding box dimensions ($[w,h]$ normalized in the range $[0,1]$) concatenated with its semantic label, which is encoded as a one-hot vector. To efficiently model document layouts using a recursive model, we first use a simple single-layer neural network to map the atomic unit bounding boxes to $n$-D vector representations (we empirically set $n=300$). 
Our recursive autoencoder network is comprised of spatial-relationship encoders (SREs) and decoders (SRDs). Each encoder and decoder is a multi-layer perceptron (MLP), formulated as:
$$x_l \ = \ \tanh \left( W^{(l)} \cdot x_{l - 1} + b^{(l)} \right).$$
We denote by $f_{W, b}(x)$ an MLP with weights $W = \{W^{(1)}, W^{(2)}, \dots\}$ and biases $b = \{b^{(1)}, b^{(2)}, \dots\}$ aggregated over all layers, operating on input $x$. Each MLP in our model has one hidden layer, and therefore, $l \in \{1,2\}$.

Our SREs may operate over either (i) a pair of leaves, or (ii) an internal node and a leaf. Regardless, we denote both node representations as $x_1$, $x_2$. The merged parent code, $y$, is calculated according to $x_1$, $x_2$ and the relative position between the two bounding boxes, denoted by $r_{{x_1}{x_2}}$. The relative position is always calculated \emph{w.r.t.} the left child (which is the internal node, when merging an internal node and a leaf node). The $i$-th SRE is formulated as:
\begin{equation}
\textit{y} = f_{W_{e_i}, b_{e_i}} ([x_1 \  x_2 \  r_{{x_1}{x_2}}]).
\end{equation}

The corresponding SRD splits the parent code \textit{y} back to its children $x'_1$ and $x'_2$ and the relative position between them $r'_{{x_1'}{x_2'}}$ (see Figure \ref{fig:pipeline}, bottom right). It uses a reverse mapping and is formulated as follows:
\begin{equation}
[x_1' \ x_2' \ r_{{x_1'}{x_2'}}'] = f_{W_{d_i}, b_{d_i}} \left( y \right).
\end{equation}
%

Each node in the hierarchy represents a feature vector, which is encoded (or decoded) by one of $c$ SREs (or SRDs). In particular, we note that since the network is recursive, the same encoder or decoder may be employed more than once for different nodes. As described in more detail below, the type of the encoder employed in each step depends on the spatial relationship between the elements in this step.

During decoding, we determine the spatial-relationship type $i$ of a node so that the corresponding decoder can be used.
To this end, we \emph{jointly} train an auxiliary node classifier to determine which SRD to apply at each recursive decoding step. This classifier is a neural network with one hidden layer that takes as input the code of a node in the hierarchy, and outputs whether the node represents a leaf or an internal node. 
In the case of an internal node, the corresponding SRD is invoked, and if it is a leaf, the code is projected back onto a labeled bounding box representation (box dimensions concatenated with a one-hot vector corresponding to the semantic category) using a non-recursive single-layer neural network.

\begin{figure}
    \centering
    \includegraphics[width=\linewidth, height = 0.6 \linewidth]{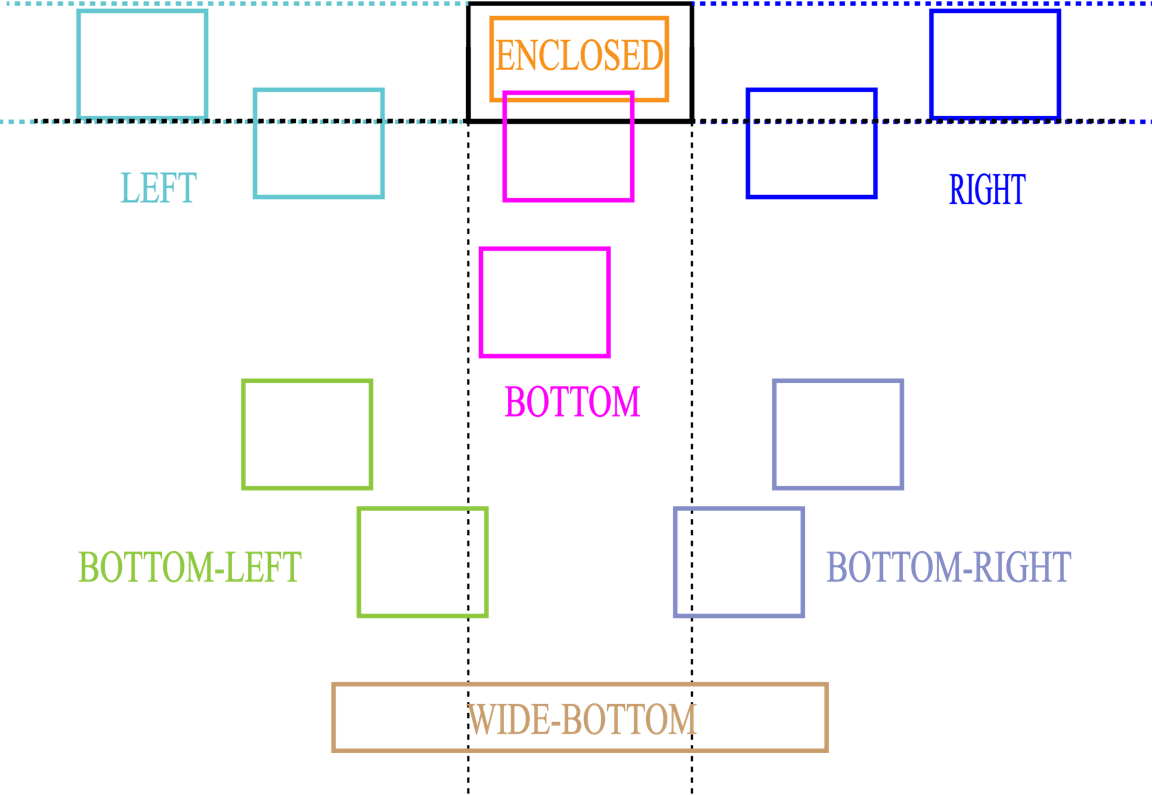}
\vspace{-16pt}
    \caption{Different types of spatial encoder/decoder pairs used in learning document layouts. The left child (or the reference box) is shown with a thick black outline. Relative positions are calculated \emph{w.r.t.} the left child. 
    }
    \label{fig:7_types_of_spatial_relations}
    \vspace{-1 em}
\end{figure}

The types of spatial relationships we consider for encoding and decoding document layouts are: right, left, bottom, bottom-left, bottom-right, enclosed and wide-bottom ($c=7$), see Figure \ref{fig:7_types_of_spatial_relations}. Note that we traverse a document from left-to-right and top-to-bottom, and therefore, we do not have to consider any kind of \emph{top} spatial relation. Please refer to the supplementary material for the full description of these spatial-relationships.

\subsection{Training details}
\label{sec:loss}
 The total training loss of our RvNN-VAE network is:
 \begin{equation}
  L_{total} = L_{leaf} + L_{pos} + L_{ce} + L_{KL}   
  \label{eq:loss}
 \end{equation}
where the first term is the leaf-level reconstruction loss:
  \begin{equation}
  L_{leaf} = \frac{1}{N}{\sum_{k=1}^{N}(x_k'- x_k)^2}.
  \end{equation}
Here, $x_k'$ and $x_k$ are the $n$-D leaf vectors at the decoder and the encoder, respectively, and N is the number of leaves.

The second term is the relative-position reconstruction loss between the bounding boxes (leaf-leaf or an internal node box and a leaf box):
 \begin{equation}
     L_{pos} = \frac{1}{N-1}{\sum_{k=1}^{N-1} (r_{{x_k'}{x_{k+1}'}}'- r_{{x_k}{x_{k+1}}})^2}
 \end{equation} 
  where  $r_{{x_k'}{x_{k+1}'}}'$ and $r_{{x_k}{x_{k+1}}}$ represent the relative position vectors at the decoder and encoder end, respectively. 
  
  The third term is a standard categorical cross-entropy loss:
    \begin{equation}
L_{ce}(\textbf{a}, i) = \log \sigma \left( \textbf{a} \right)_{i},
\end{equation}
  where $\sigma$ is the softmax function, $\textbf{a}$ is a feature vector mapped from the output of an internal (or a root) node at which the node classifier is applied, and $i \in [0, c-1]$ corresponds to the ground truth spatial-relationship type at the node.  
  

  Finally, the last term in Eq. \ref{eq:loss} is the KL-divergence loss for approximating the space of all root codes (encoder output of the RvNN-VAE):
  \begin{equation}
   L_{KL} = D_{KL}(q(z)||p(z))   
  \end{equation}
   where  $p(z)$ is the \emph{latent space} and $q(z)$ is the standard normal distribution $\mathcal{N}(0,1)$.

To train our RvNN-VAE network, we randomly initialize the weights sampled from a Gaussian distribution. To output document layouts that are more spatially balanced, we developed a few (optional) post processing steps, as explained in the supplementary material.

	\section{Evaluating Document Layouts}
To evaluate how our method performs in terms of appearance and variability, we propose 
a new combinatorial layout similarity metric we call \texttt{DocSim}.
Inspired by how the BLEU metric (bilingual evaluation understudy) for machine translation \cite{BLEU} measures sentences similarity, we aim to obtain a simple and easy-to-compute structural similarity measure between documents; one that resembles what humans perceive as similarity, yet is not too over-specified.\footnote{Generally speaking, there cannot exist a ``one-size-fits-all'' similarity metric ideal for all possible settings, as was discussed extensively regarding BLEU (see e.g.~\cite{montahaei2019jointly}). Thus, the quantitative evaluation of our paper combines \texttt{DocSim}-based comparisons with other evaluation methods, so as to try providing a complete picture of the efficacy of our approach.} We introduce our metric through the following interpretation of BLEU: consider a bipartite graph between all words $w$ in the first sentence $S$ and all words $w'$ in the second sentence $S'$, where there is an edge between $w$ and $w'$ if both represent the same word (or, say, are synonyms). The BLEU score is then calculated by computing the number of edges in a \emph{maximum matching} between these two sentences. Our metric, \texttt{DocSim}, similarly compares two given document layouts $D, D'$
as follows: to any pair of bounding boxes $B \in D$ and $B' \in D'$, we assign a \emph{weighted edge} that indicates how similar $B$ and $B'$ are in terms of shape, location, and ``role'' within the document. The final score is then calculated as the aggregated weight of the maximum (weighted) matching between the layouts $D$ and $D'$.

Formally, suppose we are given two documents, D1 and D2, each viewed as a set of bounding boxes of one or more ``types'' (examples of such types in real-world documents can be a paragraph, title, figure, and so on). Each bounding box is represented as a quadruple consisting of its minimum and maximum x and y coordinates within the document. The coordinates are \emph{normalized} to fit in the unit $1 \times 1$ square. The similarity measure between two normalized documents $D_1$ and $D_2$ is calculated in two steps: weight assignment to box pairs, and maximum weight matching among boxes. 
\begin{figure}
	\centering%
\jsubfig{\fbox{\includegraphics[height=3.65cm]{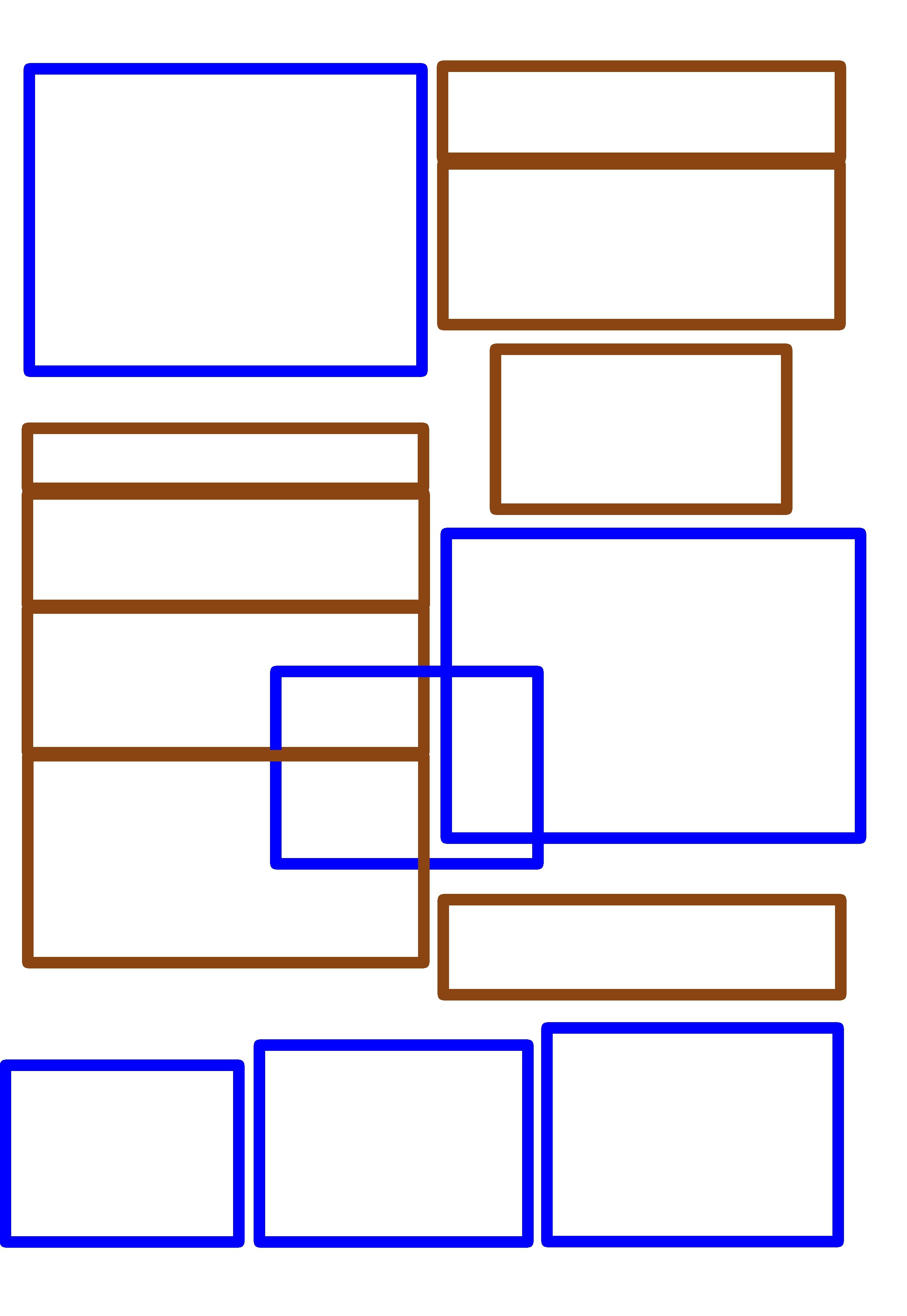}}}
	{}%
 	\hfill
\jsubfig{\fbox{\includegraphics[height=3.65cm]{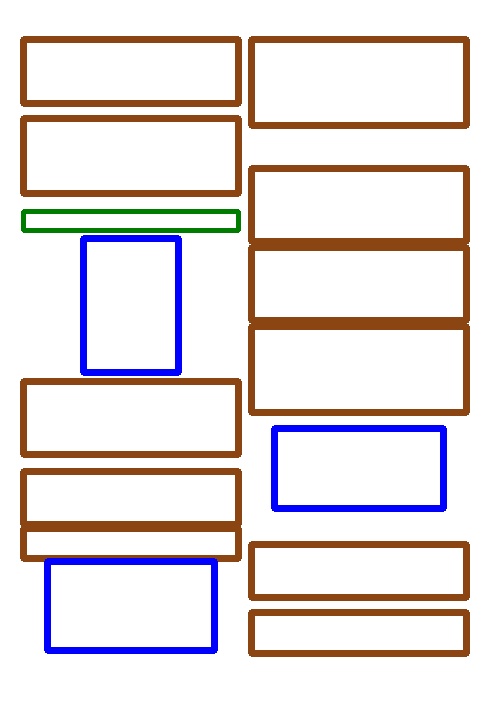}}}
	{ }%
	\hspace{2pt}
\jsubfig{\fbox{\includegraphics[height=3.65cm]{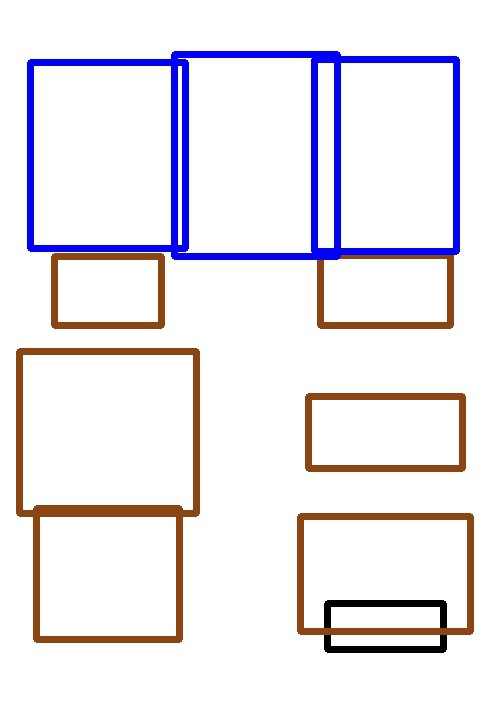}}}
	{ }%
	\\ \vspace{2pt}
\jsubfig{\fbox{\includegraphics[height=3.45cm]{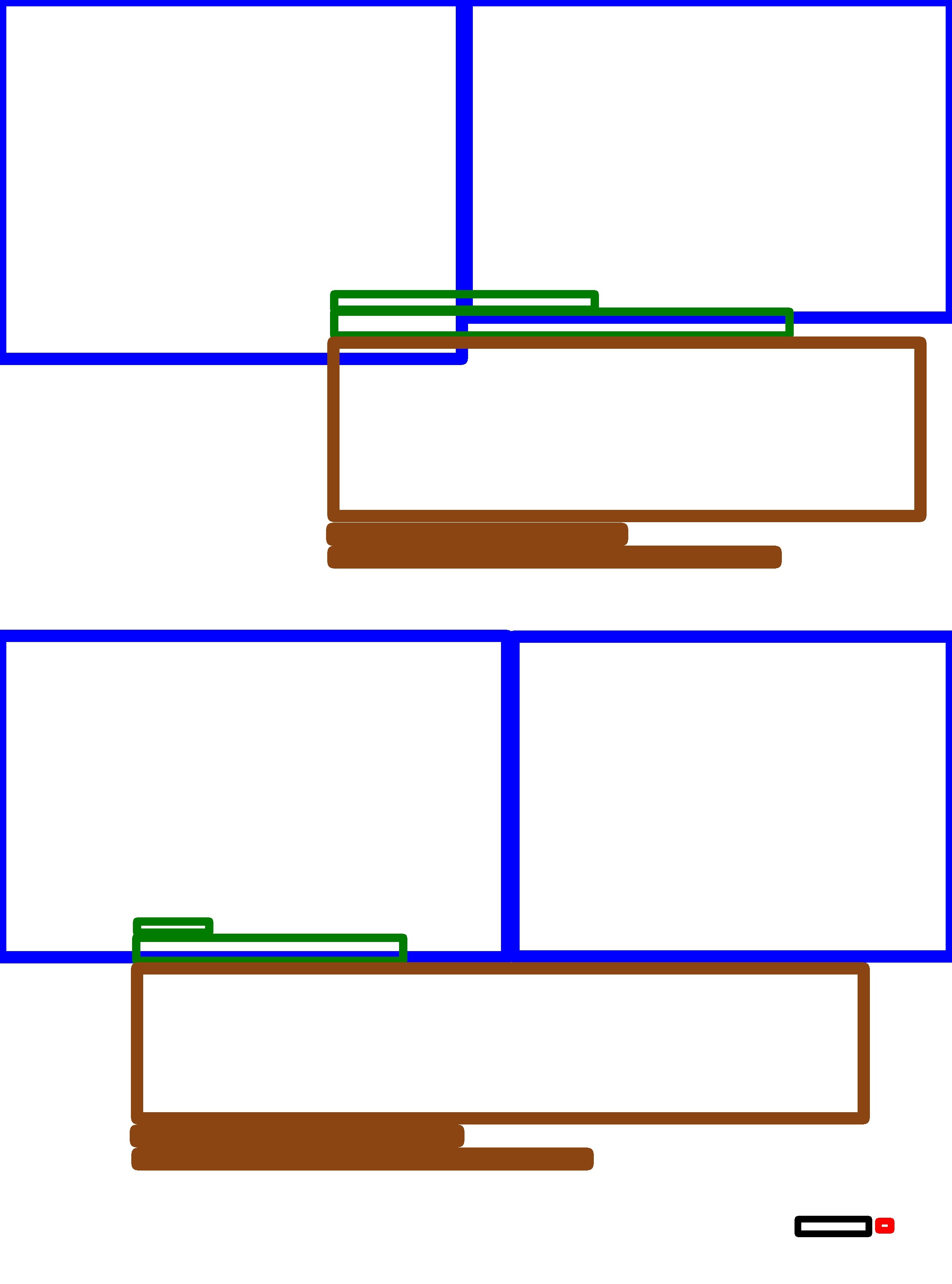}}}
	{Training sample}%
 	\hfill%
\jsubfig{\fbox{\includegraphics[height=3.60cm]{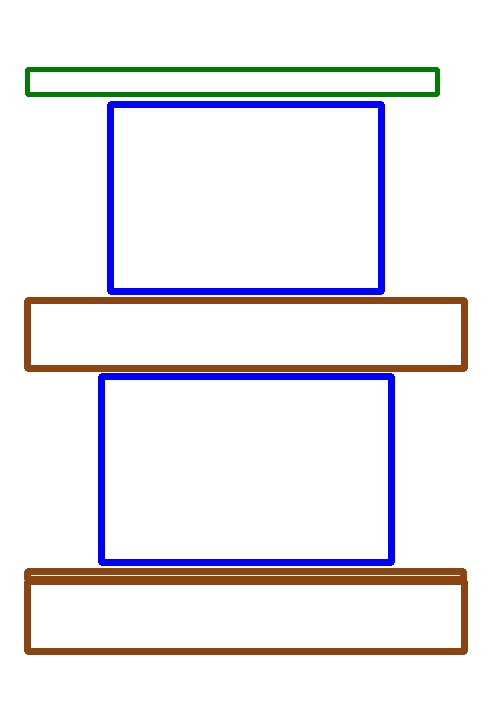}}}
	{Probabilistic \cite{yang2017learning}}%
	\hspace{4pt}
\jsubfig{\fbox{\includegraphics[height=3.60cm]{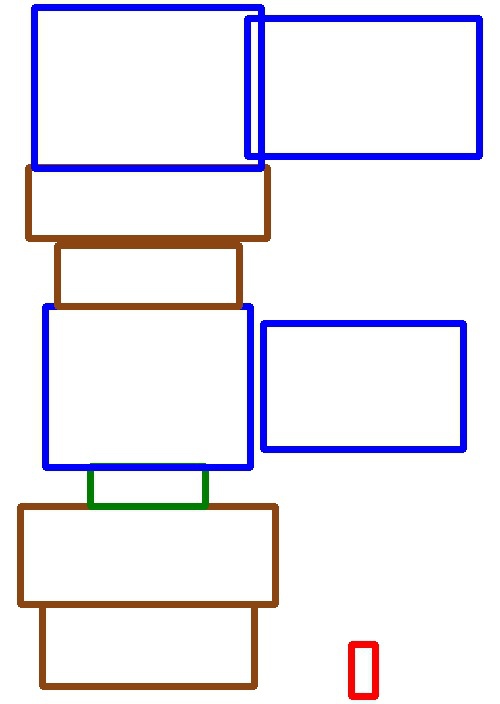}}}
	{ Ours}%
	\vspace{2pt}
	\caption{Given a document layout from ICDAR2015, we show the nearest neighbor obtained from the probabilistic approach described in \cite{yang2017learning} and the nearest neighbor using our approach.
	Color legend: \color{dark-green}{\emph{title}}, \color{saddle-brown}{\emph{Paragraph}}, \color{black} {\emph{footer}}, \color{red}{\emph{page number}}, \color{blue}{\emph{figure}}.
	}

\label{fig:compare_to_icdar}
\vspace{-1 em}
\end{figure}

    \vspace{-8pt}
    \paragraph{Assigning weights to box pairs.}
    We would like to assign weights to pairs of boxes, so that similar pairs, that are roughly co-located and have approximately the same area, will have a higher weight. In the next step, we shall use these weights to assign a maximum weight matching between boxes of $D_1$ and boxes of $D_2$; the total similarity score would simply be the total weight of the matching.
    Let $B_1$ and $B_2$ be two normalized bounding boxes, where the $x$-coordinates of box $B_i$ are denoted $a_i \leq b_i$ and its $y$-coordinated are $c_i \leq d_i$. If $B_1$ and $B_2$ have different types, then the weight between them is $W(B_1, B_2) = 0$ (this essentially means that boxes of different types cannot be matched). Otherwise, we calculate the weight as 
    \begin{equation*}
    \label{eqn:box_weight}
    W(B_1, B_2) = \alpha(B_1, B_2) 2^{ - \Delta_C(B_1, B_2) - C_S \cdot \Delta_S(B_1, B_2)}
    \end{equation*}
    where the parameters $\alpha, \Delta_C, \Delta_S$ are defined as follows:
    The \emph{location parameter} $\Delta_C(B_1, B_2)$ is the relative euclidean distance between the centers of $B_1$ and $B_2$ in the document. We wish to reduce the shared weight of $B_1$ and $B_2$ if they are far apart from each other.
    The \emph{shape difference} is $\Delta_S(B_1, B_2) = |w_1 - w_2| + |h_1 - h_2|$ where $w_i$ and $h_i$ are the width and height of $B_i$, for $i=1,2$, respectively.
    
    As larger bounding boxes have a more significant role in the ``general appearance'' of a document, we wish to assign larger weight to edges between larger boxes. Thus, we define the \emph{area factor} as $\alpha(B_1, B_2) = min(w_1 h_1, w_2 h_2)^C$, where we choose $C = 1/2$. To explain this choice, observe that changing the constant to $C = 1$ would assign almost no weight to edges between small boxes, whereas $C = 0$ strongly favors this type of edges.
    Finally, we set the \emph{shape constant} as $C_S = 2$. This means that the shape difference between two boxes plays a slightly bigger role in their weight calculation than does the location parameter.
    
    \vspace{-8pt}
    \paragraph{ Maximum weight matching among boxes.}
    Consider a bipartite graph where one part contains all boxes of $D_1$ while the other part consists of all boxes of $D_2$, and the edge weight $W(B_1, B_2)$ for $B_1 \in D_1$ and $B_2 \in D_2$ is as described above. We find a maximum weight matching $M(D_1, D_2)$ in this bipartite graph using the well-known Hungarian method \cite{Hungarian1955}. The similarity score between $D_1$ and $D_2$ is defined as 
    $$ \texttt{DocSim}(D_1, D_2) = \frac{1}{|M(D_1, D_2)|}\sum W(B_1, B_2),$$
    where the sum is over all pairs ${(B_1, B_2) \in M(D_1, D_2)}$. In the supplementary material, we provide a visualization of the matching procedure carried by \texttt{DocSim}. 
%

\begin{figure*}
	\centering%

	\jsubfig{\fbox{\includegraphics[height=3.2 cm]{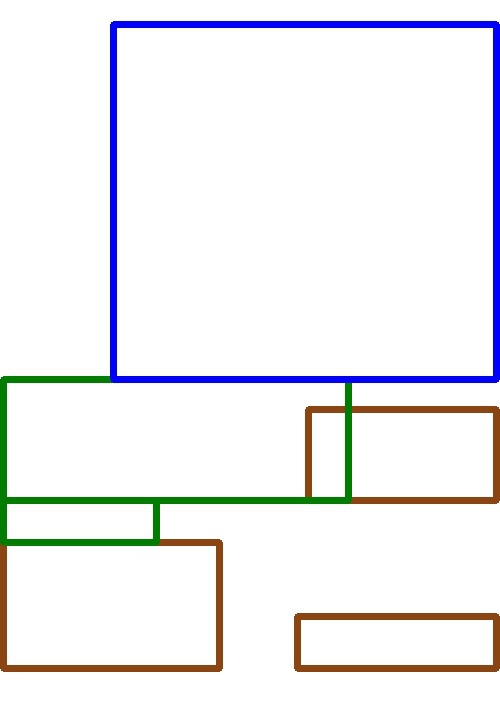}}}
	{}%
 	\hfill%
\jsubfig{\fbox{\includegraphics[height=3.2 cm]{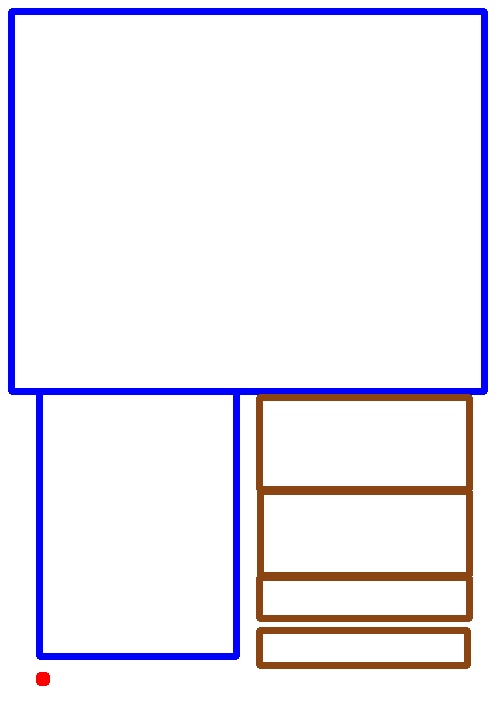}}}
	{ }%
	\hspace{1pt}
\jsubfig{\fbox{\includegraphics[height=3.2 cm]{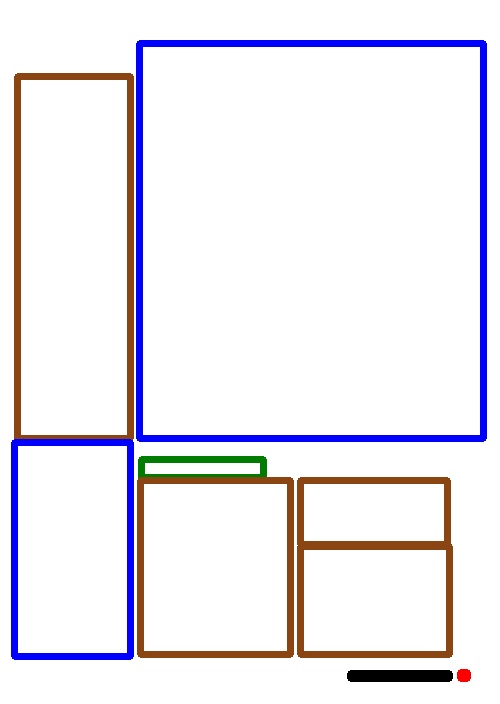}}}
	{ }%
	\hspace{1pt}
\jsubfig{\fbox{\includegraphics[height=3.2 cm]{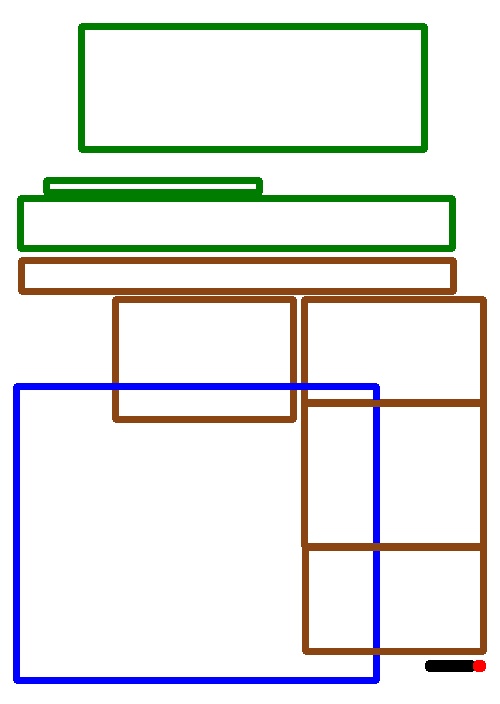}}}
	{}%
 	\hfill%
\jsubfig{\fbox{\includegraphics[height=3.2 cm]{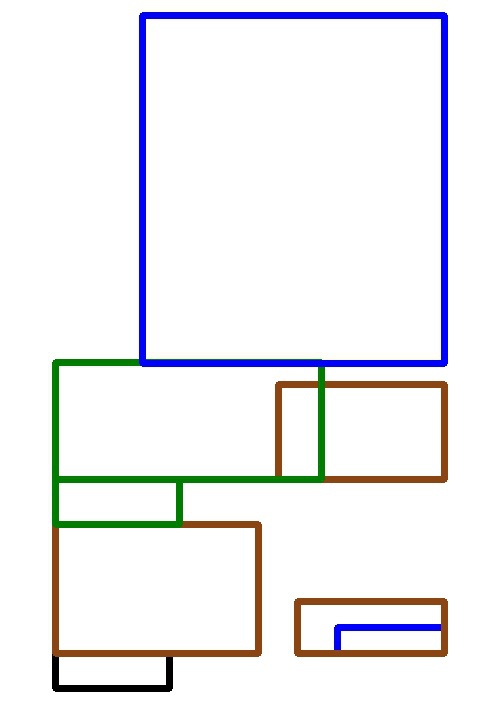}}}
	{ }%
	\hspace{1pt}
\jsubfig{\fbox{\includegraphics[height=3.2 cm]{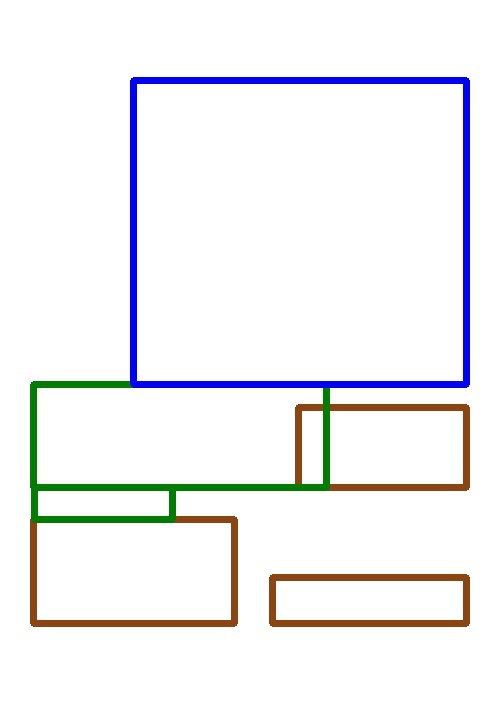}}}
	{ }%
	\hspace{1pt}
\jsubfig{\fbox{\includegraphics[height=3.2 cm]{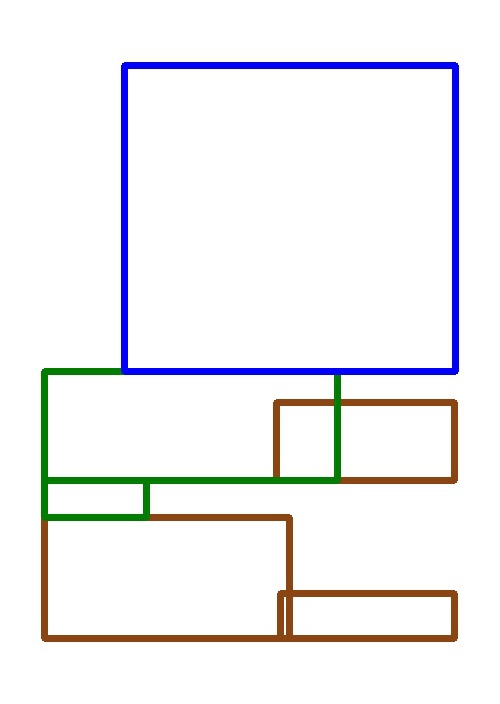}}}
	{ }%
\\
	\centering%
\jsubfig{\fbox{\includegraphics[height=3.2 cm]{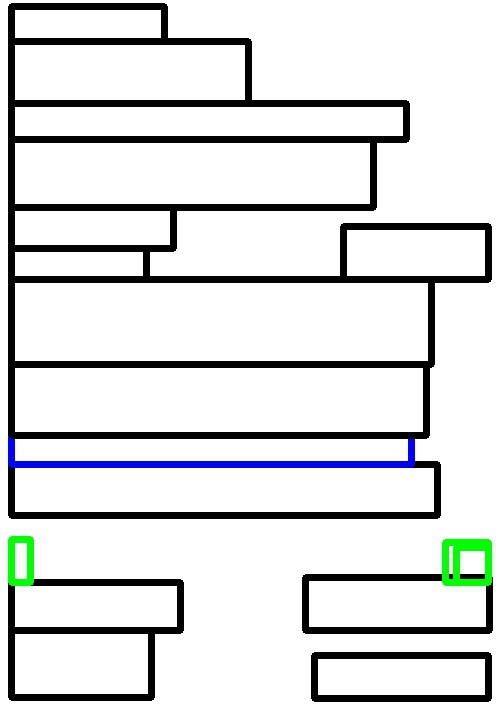}}}
	{}%
 	\hfill%
\jsubfig{\fbox{\includegraphics[height=3.2 cm]{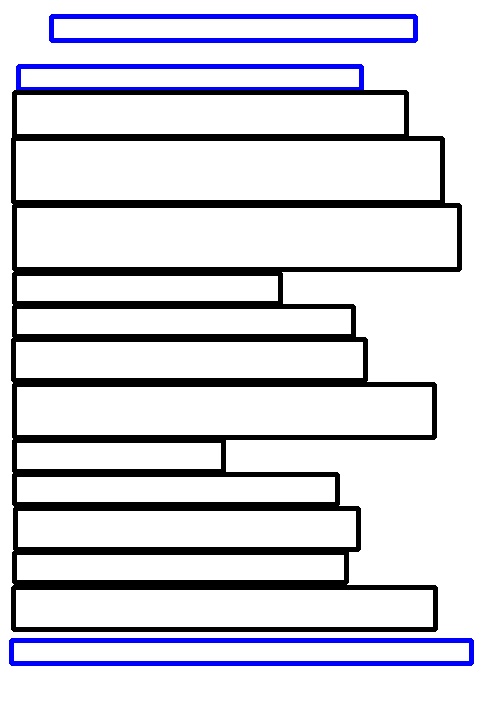}}}
	{ }%
	\hspace{1pt}
\jsubfig{\fbox{\includegraphics[height=3.2 cm]{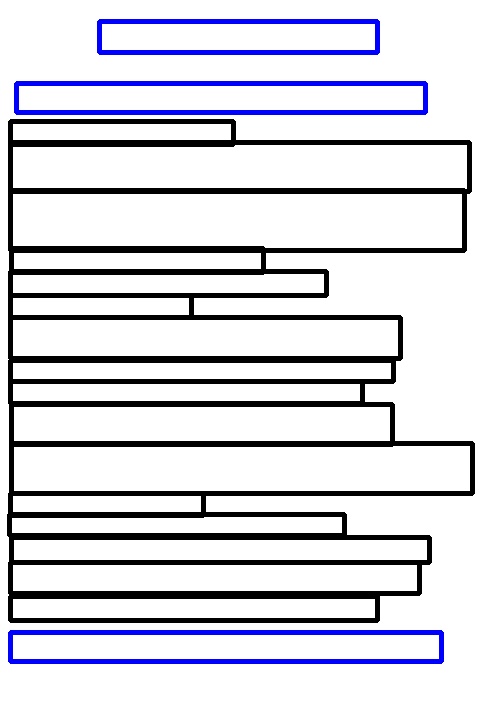}}}
	{ }%
	\hspace{1pt}
\jsubfig{\fbox{\includegraphics[height=3.2 cm]{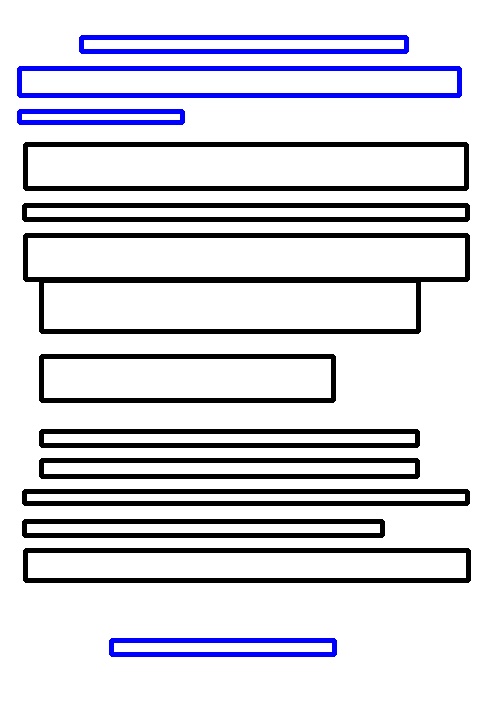}}}
	{}%
 	\hfill%
\jsubfig{\fbox{\includegraphics[height=3.2 cm]{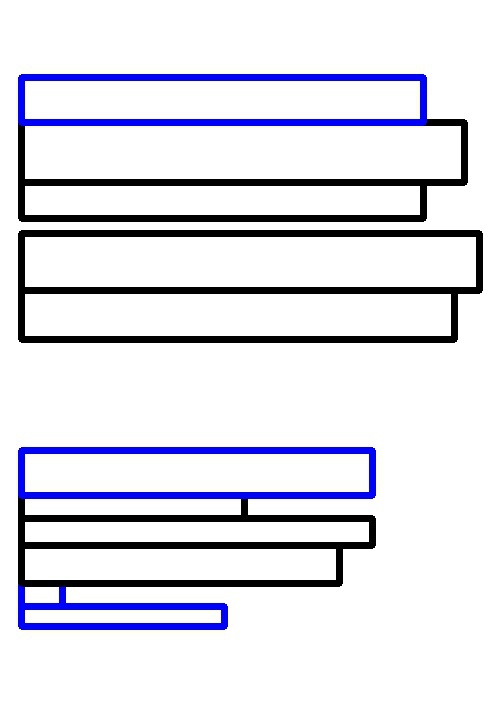}}}
	{ }%
	\hspace{1pt}
\jsubfig{\fbox{\includegraphics[height=3.2 cm]{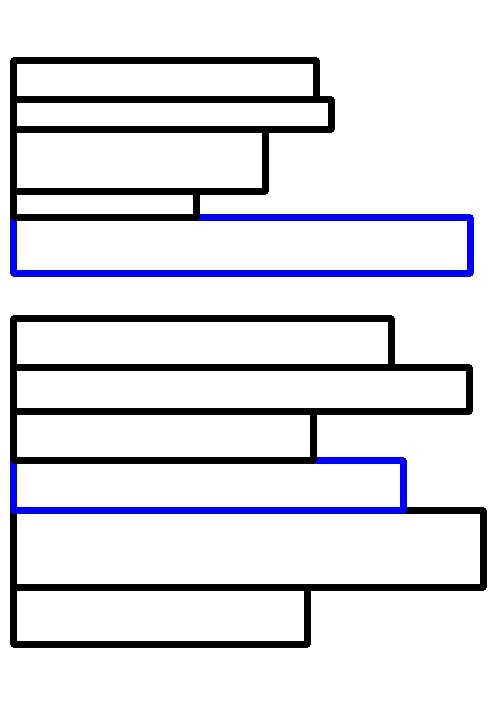}}}
	{ }%
	\hspace{1pt}
\jsubfig{\fbox{\includegraphics[height=3.2 cm]{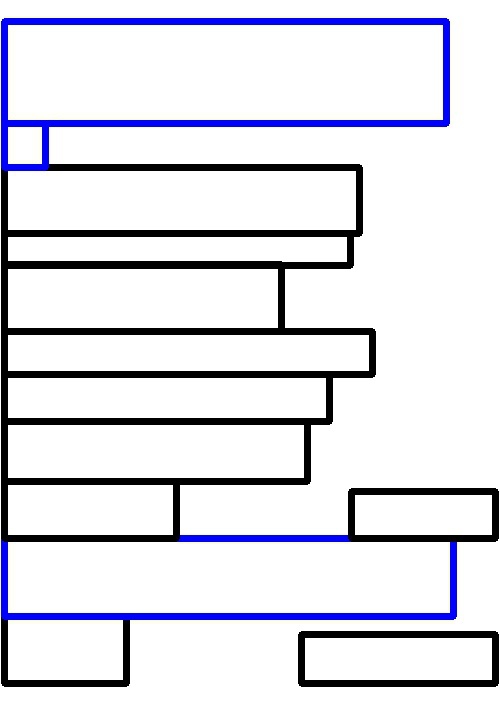}}}
	{ }%
	\\
\ignorethis{
	\jsubfig{\fbox{\includegraphics[height=3.2 cm]{Figures/KV/2329/icdar_img.jpg}}}
	{}%
 	\hfill%
\jsubfig{\fbox{\includegraphics[height=3.2 cm]{Figures/KV/2329/gen_Set_img_0_train.jpg}}}
	{ }%
	\hspace{1pt}
\jsubfig{\fbox{\includegraphics[ height=3.2 cm]{Figures/KV/2329/gen_Set_img_1_train.jpg}}}
	{ }%
	\hspace{1pt}
\jsubfig{\fbox{\includegraphics[height=3.2 cm]{Figures/KV/2329/gen_Set_img_2_train.jpg}}}
	{}%
 	\hfill%
\jsubfig{\fbox{\includegraphics[height=3.2 cm]{Figures/KV/2329/gen_Set_img_0.jpg}}}
	{ }%
	\hspace{1pt}
\jsubfig{\fbox{\includegraphics[height=3.2 cm]{Figures/KV/2329/gen_Set_img_1.jpg}}}
	{ }%
	\hspace{1pt}
\jsubfig{\fbox{\includegraphics[height=3.2 cm]{Figures/KV/2329/gen_Set_img_2.jpg}}}

	{ }%
}
	Generated sample \hspace{50pt} Nearest neighbors in train set \hspace{70pt} Nearest neighbors in generated set 
	\hspace{33pt}
	\vspace{1pt}
	\caption{Given a document layout generated by our approach, we retrieve three closest layouts from the training set (ICDAR2015 in the top row and US in the bottom row) and three closest from our generated set. \color{darkelectricblue}{Color legend (ICDAR2015): see Figure \ref{fig:compare_to_icdar}.}  \color{black}{Color legend (US):} \color{light-green}{\emph{title}}, \color{blue}{\emph{paragraph}}, \color{black}{\emph{key-value}.}}
\label{fig:3closest_KV}
\end{figure*}

\ignorethis{
	\\
	\jsubfig{\fbox{\includegraphics[height=3.2 cm]{Figures/KV/448/icdar_img.jpg}}}
	{}%
 	\hfill%
\jsubfig{\fbox{\includegraphics[height=3.2 cm]{Figures/KV/448/gen_Set_img_0_train.jpg}}}
	{ }%
	\hspace{1pt}
\jsubfig{\fbox{\includegraphics[ height=3.2 cm]{Figures/KV/448/gen_Set_img_1_train.jpg}}}
	{ }%
	\hspace{1pt}
\jsubfig{\fbox{\includegraphics[height=3.2 cm]{Figures/KV/448/gen_Set_img_2_train.jpg}}}
	{}%
 	\hfill%
\jsubfig{\fbox{\includegraphics[height=3.2 cm]{Figures/KV/448/gen_Set_img_0.jpg}}}
	{ }%
	\hspace{1pt}
\jsubfig{\fbox{\includegraphics[height=3.2 cm]{Figures/KV/448/gen_Set_img_1.jpg}}}
	{ }%
	\hspace{1pt}
\jsubfig{\fbox{\includegraphics[height=3.2 cm]{Figures/KV/448/gen_Set_img_2.jpg}}}
	{ }%

}

\section{Results and Evaluation}
\label{sec:results}

To assess our layout generation method, we conducted several sets of experiments, aiming at understanding whether the generated layouts are highly variable and also \emph{visually}-similar to the training documents. 
We also demonstrate their usefulness as training data for document analysis tasks. In the supplementary material, we provide a detailed ablation analysis explaining our design choices in terms of the number of SRE/SRDs. We evaluate our RvNN-VAE framework on the following two datasets.
%
%
    \vspace{-10 pt}
    \paragraph{ICDAR2015 Dataset.} We use the publicly available ICDAR2015 \cite{icdar2015} dataset, containing $478$ documents that are themed along the lines of magazine-articles. 
    For these documents, we consider the following semantic categories:
    \emph{title}, \emph{paragraph}, \emph{footer}, \emph{page number}, and \emph{figure}.
    
    \vspace{-10 pt}
    \paragraph{User-Solicited (US) Dataset.} We assembled a dataset of $2036$ documents that solicit user-information (tax forms, banking applications, etc.). Such documents typically exhibit a highly complex structure and a large number of atomic elements.
    These characteristics present an interesting challenge for generative models producing document layouts. For these types of documents, we consider the following semantic categories: \emph{key-value}, \emph{title}, and \emph{paragraph}. Key-value boxes are regions with a single question (key) that the user must answer/address (value). As the dataset we collected captures unfilled documents, the key-value box contains regions that should be filled out by the user. 
    We semantically annotated all the categories 
    using Amazon Mechanical Turk (AMT).
   
  \vspace{3mm} 
  \textbf{Training:} We use the PyTorch framework \cite{paszke2017automatic}, with a batch size of 128 and a learning rate of $3*10^{-4}$. On average, the number of semantically annotated bounding boxes is 27.73 (min=13, max=45) in the US \emph{training} set and 17.61 (min=3, max=75) for ICDAR2015 \emph{training} set. As is shown in the two rightmost columns of Table \ref{tbl:ours_vs_layoutGAN}, the statistics on our generated data are similar. Training takes close to 24 hours on the US dataset and around 10 hours on the ICDAR2015 dataset, on an NVIDIA GTX 1080 Ti GPU.
    

\subsection{Quantitative evaluation}
We use our proposed similarity metric, \texttt{DocSim}, to quantitatively evaluate our layout generation approach. 
To measure resemblance of our generated document layouts to the latent distribution of document layouts from which the training data is sampled from, we iterate over training-set and test-set, and for each document in these sets, we find the nearest neighbor in our generated layouts. To this end, the nearest neighbor of a document $D$ is the document $D'$ which \emph{maximizes} the score \texttt{DocSim}($D$, $D'$), and correspondingly, the similarity score that $D$ has with respect to a dataset $\mathcal{D}$ is defined as $\max_{D' \in \mathcal{D}} \texttt{DocSim}(D, D')$. In our nearest neighbors experiments, we filter out documents $D'$ whose number of boxes from any category is more than 3 higher or lower (before overlap removal) than that of $D$.
\vspace{-10pt}
\paragraph{On the ICDAR2015 dataset.}
As a baseline, we obtain synthetic layouts using the probabilistic approach described in \cite{yang2017learning}, using their publicly available implementation.
Notably, the main focus of \cite{yang2017learning} is semantic segmentation of documents, and their probabilistic layout synthesis method (which outputs one-, two- and three-column documents) is developed as a helper for their main learning task. 

In the probabilistic synthesis method of \cite{yang2017learning}, labeled boxes are sampled according to a pre-defined distribution (e.g., a \emph{paragraph} is selected with probability $q$). 
We obtain a collection $\mathcal{P}$ of 5k layouts using the probabilistic scheme of \cite{yang2017learning}; layouts are synthesized with the \emph{title}, \emph{paragraph} and \emph{figure} classes, selected at probability $0.1$, $0.7$ and $0.2$, respectively. 
Similarly, we obtain a collection $\mathcal{G}$ of 5k layouts generated by our RVNN-VAE framework, where we use a training set $\mathcal{T}$ of 400 documents from ICDAR2015. The collection $\mathcal{T}'$ of all remaining 78 documents from ICDAR2015 is considered our test set.

We experiment by comparing the baseline collection $\mathcal{P}$ with our collection $\mathcal{G}$ in terms of how well they capture the latent document layout space, where the evaluation uses our \texttt{DocSim} score. First, we run the following: for any training document $T \in \mathcal{T}$, we pick $G_T \in \mathcal{G}$ to be the generated document from our collection which maximizes $\texttt{DocSim}(T, G)$ among all $G \in \mathcal{G}$, and similarly $P_T \in \mathcal{P}$ as the document from the probabilistically synthesized collection which maximizes $\texttt{DocSim}(T, P)$ among all $P \in \mathcal{P}$. The similarity score between $\mathcal{T}$ and $\mathcal{G}$ is then calculated as the average of $\texttt{DocSim}(T, G_T)$ over all $T \in \mathcal{T}$; the similarity score between $\mathcal{T}$ and $\mathcal{P}$ is computed analogously using $\texttt{DocSim}(T, P_T)$ for all $T \in \mathcal{T}$. Finally, we repeat the above experiment, replacing the training set $\mathcal{T}$ with the test set $\mathcal{T}'$. 



The scores, given in Table \ref{tbl:prima_sim_scores}, demonstrate that our \emph{learned} document layouts are more structurally-similar to samples in the ICDAR2015 dataset, suggesting that our network is able to meaningfully learn the latent distribution of document layouts on which it was trained.

\begin{table}
  \begin{center}
  \setlength{\tabcolsep}{0.37em} 
    \begin{tabular}{ lccccccc }
      \toprule
       	Measure	& \thead{Real \\ \cite{icdar2015} } & \thead{Probabilistic \\ \cite{yang2017learning}}&
            \thead{Generated \\ (Ours)}\\
        \midrule
       	Overlap (\%) & 2.3 & 0 & 1.9 \\
        Alignment (\%) & 17.8  & 9.6 & 18.2 \\
      \bottomrule
    \end{tabular}
  \end{center}
 \vspace{-8pt}
  \caption{Spatial analysis of document layouts. Following \cite{li2019layoutgan}, we use overlap index and alignment index of semantic entities as another measure to evaluate our layouts.}
  \label{tbl:spatial_analysis_measure_}
  \vspace{-1 em}
\end{table}
In addition, we perform a quantitative analysis using the overlap and alignment indices, following the evaluation in Li et al.~\cite{li2019layoutgan}. Overlap index is the percentage of total overlapping area among any two bounding boxes inside the whole page. The second metric, alignment index, is calculated by finding the minimum standard deviation of either left or center coordinates of all bounding boxes. Table \ref{tbl:spatial_analysis_measure_} shows the percentage of overlap index and alignment index for the real ICDAR2015 layouts \cite{icdar2015}, probabilistic layouts \cite{yang2017learning} and our generated layouts. As illustrated in the table, our results are very much comparable to those of the training data, demonstrating that our solution captures these metrics well (and does much better than the probabilistic layouts). In the supplementary, we also show the distribution of box centers and their dimensions in a layout.

\vspace{-10pt}
\paragraph{On the US dataset.}
As we are not aware of prior works that address these types of documents, we do not have a baseline method to compare to. We can, however, investigate the learning ability of our network on this dataset, which contains a relatively large number of documents (2036).  
Therefore, aside from training our network on the full dataset, we also use smaller subsets of training samples. As the entire US dataset is highly-variable, we compute our similarity score for every pair of document layouts in the entire US dataset and cluster the dataset into five groups (using spectral clustering). We then train our network on clusters that contain at least 500 documents, using a 80-20 train and test split, and generate 2$K$ document layouts for each cluster.

We then compare the similarity scores obtained by training on the entire US dataset against the scores obtained on the US clusters (averaging over all cluster scores). Interestingly, the scores of the train/test sets are virtually almost identical (with a slight score advantage of $0.002$ to $0.003$ for the entire US dataset, which is a $2-3\%$ advantage).
This suggests that our approach does not require a large amount of data to match the latent space of the training set reasonably well; Indeed, as indicated by the relatively similar scores, the models trained on the clusters capture the latent space of the training set roughly as good as the model that was trained on the full set.
%
In Figure \ref{fig:3closest_KV}, we show the three closest document layouts from the training set 
to a randomly selected layout sample 
generated using our approach. As the middle three columns demonstrate, the three closest training samples bear some resemblance to our generated layouts, but they are not the same, further validating the novelty of the generated samples. The rightmost column, depicting the nearest neighbors in the generated set, illustrates the variations in the generated results. See the supplementary material for more results.
\begin{table}
  \begin{center}
  \setlength{\tabcolsep}{0.37em} 
    \begin{tabular}{ lccccccc }
      \toprule
       	ICDAR \cite{icdar2015}	& \cite{yang2017learning} & {Ours} \\
        \midrule
       	Train (400) & 0.123 & \bf{0.147} \\
       	Test (78) & 0.118 & \bf{0.146} \\
      \bottomrule
    \end{tabular}
  \end{center}
 \vspace{-5pt}
  \caption{Comparing our approach to the probabilistic approach from \protect\cite{yang2017learning}, in terms of similarity to the latent distribution of the dataset (divided into train and test).}
  \label{tbl:prima_sim_scores}
  \vspace{-0.5 em}
\end{table}

\subsection{Data augmentation for detection tasks}
To demonstrate the utility of our generated layouts, we perform a standard detection task on documents and augment the training data with generated documents whose layouts are produced by our method. We train Mask R-CNN \cite{he2017mask}, a popular object detection and segmentation network, on the ICDAR2015 dataset and evaluate the results obtained with and without performing data augmentation.

\label{subsec:augment_training}

\begin{table}
\centering
\ra{0.96}
\setlength{\tabcolsep}{2.6pt}
\begin{tabular}{lccccccllllll}
\toprule
\multicolumn{1}{c}{} &\phantom{} & \multicolumn{3}{c}{Box IoU} & \phantom{} & \multicolumn{3}{c}{Mask IoU}                             \\
Dataset              && \multicolumn{1}{l}{$AP$} & \multicolumn{1}{l}{$AP_{50}$} & \multicolumn{1}{l}{$AP_{75}$} &&  $AP$  & $AP_{50}$ & $AP_{75}$  \\ 

\midrule
\cite{icdar2015}    && 0.609     & 0.743   & 0.675  && 0.612 & 0.737     & 0.675      \\
\cite{icdar2015}+5k (aug.)                  && 0.611                    & 0.728                         & 0.663      && 0.617 & 0.722     & 0.669      \\
\cite{icdar2015}+5k (\cite{yang2017learning})                  && 0.605                    & 0.753                         & 0.676      && 0.612 & 0.750     & 0.665       \\
\cite{icdar2015}+5k (ours)           && \textbf{0.634}                    & \textbf{0.770}                         & \textbf{0.702}                                             &&\textbf{0.644} & \textbf{0.769}     & \textbf{0.700}      \\ 
\bottomrule
\end{tabular}
\vspace{-3pt}
  \caption{Enhancing detection and segmentation performance on the ICDAR2015 \cite{icdar2015} dataset using either data augmentations (second row), synthetic samples with probabilistic layouts (third row) or our learned layouts (bottom row).}
  \label{tbl:ap_all}
  \vspace{-13pt}
\end{table}


To generate training samples for Mask R-CNN, we inject content to our generated layouts (trained on $400$ documents from the ICDAR2015 dataset). To do so, we scrape both text and images from Wikipedia. We also synthesize training samples using the probabilistic approach described in \cite{yang2017learning}, and compare our results to the ones obtained by augmenting the dataset with their documents. The content in both cases is sampled from the same scraped data, thus the only difference is in the layouts. \rev{Furthermore, we compare our results to a standard augmentation technique, which uses photometric and geometric augmentations to enrich the ICDAR2015 dataset (see the supplementary material for a few augmented samples). }
%
%
%
In Table \ref{tbl:ap_all}, we compare the bounding box detections and the segmentation results obtained by training on the different datasets. For both types of results (box/mask), we report the average precision ($AP$) scores averaged over IoU thresholds and at specific IoU values ($AP_{50}$, $AP_{75}$). The reported results are over the remaining $78$ documents, which we do not train on.
As the table demonstrates, our generated layouts consistently improve detection and segmentation IoU scores (by at least $3\%$). In comparison, scores obtained with documents synthesized using the probabilistic approach or using regular augmentation techniques are almost identical to the scores obtained on the dataset without any augmentations.
The improved performance illustrates the vast importance of highly variable layout in generating meaningful synthetic data, validating that our technique successfully learns a layout distribution which is similar to the input dataset. 
\vspace{-3 pt}
\begin{table}
  \vspace{-5pt}
  \begin{center}
  \setlength{\tabcolsep}{0.37em} 
    \begin{tabular}{ lccccccc }
      \toprule
       	Method	& \thead{\#Training \\ samples} & 
            \thead{\#Semantic \\ categories}&
            \thead{\#Boxes \\ Avg.}&
            \thead{\#Boxes \\ Max}\\
        \midrule
       	\cite{li2019layoutgan} & 25000 & 6 & - & 9 \\
       	Ours (on \cite{icdar2015})  & 400  &5& 17.06 & 74\\
       	Ours (on \textsc{US}) 	& 560 & 3	& 28.27 & 45\\
        
      \bottomrule
    \end{tabular}
  \end{center}
 \vspace{-8pt}
  \caption{Comparison to previous work in terms of number of samples used for \emph{training}, number of semantic categories in the \emph{training} set, and average number of boxes per \emph{generated} document.}
  \label{tbl:ours_vs_layoutGAN}
  \vspace{-1 em}
\end{table}
\subsection{Comparison to prior work}
\vspace{-1 pt}
To the best of our knowledge, LayoutGAN \cite{li2019layoutgan} is the only prior work for our context. 
For the lack of publicly available code and dataset from \cite{li2019layoutgan}, we perform a quantitative comparison on  methodological statistics and present them in Table \ref{tbl:ours_vs_layoutGAN}, and as was done in \cite{li2019layoutgan}, we use the overlap and alignment metrics (as described before) to compare between real layouts, our generated ones, and probabilistic layouts (see Table \ref{tbl:spatial_analysis_measure_}).
	\section{Conclusions}
\label{sec:conclusions}
\vspace{-2 pt}
In this work, we have presented a new method for generating synthetic layouts for 2D documents, involving a recursive neural network coupled with a variational autoencoder.
We also introduced a metric for measuring document similarity, \texttt{DocSim}, and used this metric to show the novelty and diversity of our generated layouts. 

\ignorethis{
Finally, we have validated that our generated layouts may be useful for learning: specifically, that augmenting real annotated training data (in our case, the ICDAR2015 \cite{icdar2015} dataset) with our synthetic data boosts the performance on standard detection tasks.} 

There are several limitations to our approach. First, while our approach can generate highly variable layouts with dozens of elements, we are not yet able to generate highly complex layouts (e.g., the US tax form 1040), and it will be very interesting to understand how to reliably represent and generate such layouts. Second, our generated layouts may contain undesirable artifacts, such as misalignment and box overlaps. 
%
We addressed these artifacts using simple heuristics, but perhaps a more systematic solution would be to couple the current framework with a GAN, which will encourage the generated layouts to be more visually similar to the training samples. 

In the future, it will be interesting to complement our layout generation approach with a suitable way to generate high quality semantic content that ``makes sense'' in view of the layout. Additionally, while our network does not require a huge amount of annotated data, it remains to be seen if there is a way to devise layout generation methods that require even less annotated training data, perhaps one-shot or few-shot approaches to generate plausible and ``similarly looking'' layouts. Finally, while recursive neural networks were shown (here and in previous works) to be useful for generating ``human-made'' hierarchical structures, like documents and indoor scenes, can they be incorporated for generating highly structured \emph{natural} scenes?

\ignorethis{
.....approach with a suitable way to generate high quality semantic content that ``makes sense'' in view of the layout. To this end, an interesting recent work of Liu et al.~\cite{DesignSemantics2018} on design semantics for mobile apps  shows that the semantic roles of different parts of the document can be deduced (with high level of granularity) from the layout, using deep learning based approaches.}

\ignorethis{
\begin{figure}
    \centering
    \includegraphics[width=\linewidth, height = 0.6 \linewidth]{Figures/spatial_relations_oneBox.jpg}
    \caption{The different types of spatial encoder/decoder pairs used in learning document-structures. Relative positions are calculated w.r.t the left child. The left child (or the reference box) is shown with a thick black outline. Since we traverse a document from left-to-right and top-to-bottom, we do not have to consider any kind of \emph{top} spatial relation.
    \todo{Another option for spatial relationships figure}}
    \label{fig:7_types_of_spatial_relations_}
\end{figure}
}
	
	{\small
		\bibliographystyle{ieee_fullname}
		\bibliography{7-References}
	}
	
\end{document}